\documentclass{article}
\usepackage{arxiv}

\usepackage{amsmath,amssymb,amsfonts}
\usepackage{algorithmic}
\usepackage{graphicx}
\usepackage[export]{adjustbox}
\usepackage{textcomp}
\usepackage{xcolor}
\usepackage{biblatex}
\usepackage{booktabs}
\usepackage{multirow}
\usepackage{subcaption}
\usepackage{svg}

\usepackage{hyperref}
\usepackage{cleveref}
\def\BibTeX{{\rm B\kern-.05em{\sc i\kern-.025em b}\kern-.08em
    T\kern-.1667em\lower.7ex\hbox{E}\kern-.125emX}}
\renewcommand{\shorttitle}{Optimized Drug Design using Multi-Objective Evolutionary Algorithms with SELFIES}

\title{
Optimized Drug Design using Multi-Objective Evolutionary Algorithms with SELFIES
}

\author{ 
{Tomoya Hömberg}\\
Department of Computational Intelligence\\
Otto-von-Guericke University\\
Magdeburg, Germany\\
\And
{Sanaz Mostaghim}\\
Department of Computational Intelligence\\
Otto-von-Guericke University\\
Magdeburg, Germany\\
\And
{Satoru Hiwa}\\
Faculty of Life and Medical Sciences\\
Doshisha University\\
Kyoto, Japan\\
\And
{Tomoyuki Hiroyasu}\\
Faculty of Life and Medical Sciences\\
Doshisha University\\
Kyoto, Japan\\
\texttt{tomo@is.doshisha.ac.jp} 
}

\date{}

\begin{document}
\maketitle

\begin{abstract}
Computer aided drug design is a promising approach to reduce the tremendous costs, i.e. time and resources, for developing new medicinal drugs. It finds application in aiding the traversal of the vast chemical space of potentially useful compounds. In this paper, we deploy multi-objective evolutionary algorithms, namely NSGA-II, NSGA-III, and MOEA/D, for this purpose. 
At the same time, we used the SELFIES string representation method.
In addition to the QED and SA score, we optimize compounds using the GuacaMol benchmark multi-objective task sets. Our results indicate that all three algorithms show converging behavior and successfully optimize the defined criteria whilst differing mainly in the number of potential solutions found. We observe that novel and promising candidates for synthesis are discovered among obtained compounds in the Pareto-sets.
\end{abstract}


\section{Introduction}
The field of drug discovery aims to invent useful new medicines for many reasons. New medicine can facilitate more effective and affordable treatment of known diseases, make previously incurable or unknown conditions curable, and improve health in general. A major drive behind said research is that a plethora of undiscovered drugs exist that have the potential to aid in the creation of such novel medications. This search space is known as the chemical space and is so unimaginably vast that it is practically impossible to explore exhaustively \cite{29}. Instead, a more narrow and targeted search is required.

A forthright strategy is, to only consider the subspace of chemical compounds likely to be suitable drugs \cite{24}. A qualitative assessment concept to identify such compounds is drug-likeness. The essential idea behind drug-likeness is to examine whether a compound shares desired properties with well-known, established drugs. For instance, simple features such as molecular weight or more complex metrics, like lipophilicity or synthesizability \cite{33}. Nonetheless, even this subspace is estimated to an enormous size of around $10^{33}$ \cite{30} and it remains a cumbersome and costly task to traverse effectively. 

In the hopes of alleviating these issues, recent advances in computational methods attempt to accelerate this process through in silico experiments. This process is known as computer-aided drug design (CADD). One such approach of CADD is to design new drug candidates based on information from known active compounds. Here, we discuss the method of designing compounds that will become drug candidates. The area of artificial intelligence, especially deep learning (DL), has seen a surge in popularity, with many DL systems for drug design being published in recent years \cite{32, 12}. But DL methods have their drawbacks; they heavily rely on extensive input data, and a considerable amount of time and computational resources must be allocated to training. Furthermore, interpretability is a significant problem, especially as the best performing models are increasingly complex. 

An alternative approach to circumvent these issues is formulating and determining the issue of designing new drug candidate compounds as an optimization problem. When evaluating compounds, it is necessary to consider multiple indices such as drug-likeness and ease of synthesis. Such a situation corresponds to a multi-objective optimization problem for which various methods can be applied. For example, using Evolutionary Algorithms (EAs) \cite{21} is of particular interest. EAs are known in multi-objective optimization problems for their simplicity, ease of interpretation, and have shown to provide competitive results \cite{18}.

When applying EAs to compound design, it is necessary to convert the structure of the compounds into a computer readable format, e.g. a string representation. Although several methods exist for this conversion, the most prominent technique is the Simplified Molecular-Input Line-Entry System (SMILES) \cite{35}. However, a significant issue with SMILES is the low probability of random strings forming valid compound structures. There is no guarantee that a new string representation, created from the combination of parts of SMILES representations, will correspond to a feasible compound. This problem implies that the string representation of offspring generated through EA crossover does not always correspond to a viable structure. Consequently, the use of SMILES can result in inefficient exploration. 

On the other hand, SELF-referencing Embedded Strings (SELFIES) \cite{19}, a recent method for converting to string representation, guarantees that random strings will form valid structures. Therefore, SELFIES is expected to be effective in EA-based exploration.
 
This study's primary contribution lies in exploring the potential of Multi-Objective Evolutionary Algorithms (MOEAs) in the design of compounds as candidates for new drugs, employing the string representation method SELFIES. To achieve this objective, we applied three well-known evolutionary multi-objective optimization algorithms: NSGA-II \cite{09}, NSGA-III \cite{08}, and MOEA-D \cite{37}, and compared their respective outcomes. The objective functions utilized are a combination of objectives selected from the GuacaMol \cite{06} benchmark suite and two established metrics. This research demonstrates that using SELFIES in conjunction with MOEAs can yield a diverse set of solutions. The second significant contribution of this paper is an analysis of the compounds (solutions) obtained to investigate whether MOEAs can discover new solutions that are not present in conventional databases. This analysis reveals that MOEAs not only have the capability to find undiscovered drugs but compounds that encompass desirable properties in potential pharmaceuticals, suggesting a substantial contribution to the discovery of new medications.

\begin{figure}[t!]
\centerline{\includegraphics[width=0.45\textwidth]{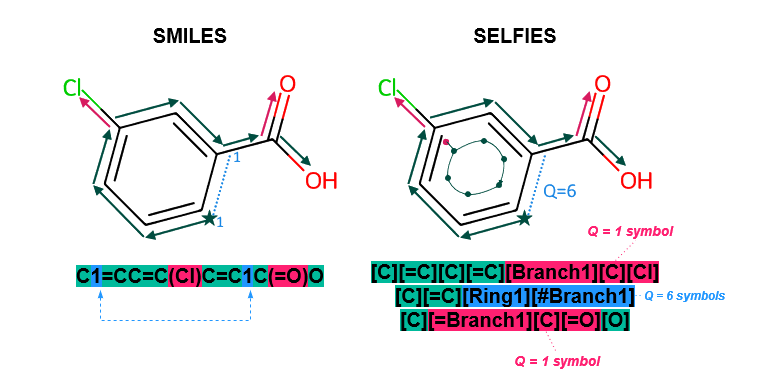}}
\caption{SMILES vs SELFIES representation. The star indicates the starting position of derivation.}
\label{fig:smilesselfies}
\end{figure}

\begin{figure*}[!ht]
    \centering
    \fbox{\includegraphics[width=0.9\textwidth]{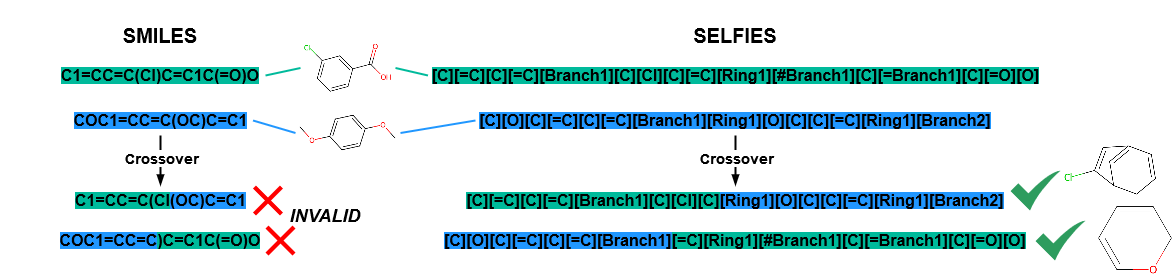}}
    \caption{SMILES vs SELFIES crossover.}
    \label{fig:crossover}
\end{figure*}

\section{Related Work and Methods}

EAs have been proven as a successful tool when applied to the problem of CADD \cite{10}. Although deep learning approaches were more prominent in CADD in recent years, as of late, EAs have seen a resurgence in activity, due to producing competitive results, despite their simplicity.

\subsection{Related Work}

Early attempts at using EAs for drug design relied on a single-objective function to accommodate all desired traits of a compound. For example in the work of Glen and Payne \cite{14}, a weighted sum that checks for satisfaction of an enumeration of many constraints handles the evaluation of compounds. Douguet, Thoreau, and Grassy follow a similar approach in their SMILES based EA \cite{11}, differing in used constraints.

Taking a leap forward to recent techniques, more sophisticated evaluation and algorithmic methods have emerged. Yoshikawa et al. \cite{36} use a context-free grammar of SMILES to evolve compounds, while measuring diversity and drug-likeness. They report successful generation of diverse compounds with high scores that compete with other DL systems.

Kwon and Lee \cite{22} present MolFinder, a SMILES based EA using the Conformational Space Annealing algorithm. Their results outperform machine learning approaches, whilst also being computationally more efficient. However, their method displays the hindrance in SMILES, as their genetic operators rely on repair mechanisms as well as trial and error to produce valid offspring.

Jensen \cite{18} implemented a graph-based Genetic Algorithm with Monte Carlo Tree Search. It matches machine learning based techniques in compound quality and exceeds in computing speed. Based on his findings, Jensen makes a strong argument in favour of EAs regarding their simplicity, and in a follow up work, Henault, Rasmussen, and Jensen \cite{16} highlight this argument by emphasizing the ability of EAs, for effective chemical space exploration. Another graph-based approach is EvoMol by Leguy et al. \cite{23}. They report remarkable performance for evolving compounds for drug-likeness, but also for a use case relating to electronic properties, optimizing HOMO and LUMO energies. Furthermore, they built a visualization of the algorithms exploration, providing more insight and interpretability.

STONED is an algorithm by Nigam et al. \cite{28} that specializes in the traversal of chemical space by applying mutations to SELFIES strings. They too achieve notable results both in drug-likeness and photovoltaics. They stress the advantage of not relying on previous data and the overall straightforwardness of their algorithm.

Lastly, Verhellen \cite{34} uses MOEAs to evolve graphs. Specifically, he compares NSGA-II and NSGA-III to a baseline single-objective EA. The solutions are evaluated via dominated hypervolume and the geometric mean, and results show that both NSGA-II and III noticeably outperform the baseline. Moreover, he calculates the internal similarity to gauge diversity within the final populations. Verhellen's work is closest to our work, in the sense that used algorithms, objectives, and evaluation measures overlap.

\subsection{Molecular Structure Representation}

In the identification of 'drug-like' compounds using evolutionary computation, converting the phenotypic representation of a compound into a genotypic string format is essential. Two primary methods are widely utilized to facilitate this transformation: SMILES (Simplified Molecular-Input Line-Entry System) and SELF-referencing Embedded String (SELFIES). SMILES offers a comprehensive approach for representing molecules as strings; however, many generated strings can result in chemically invalid outcomes. Conversely, SELFIES ensures that every combination of symbols maps to a chemically valid graph, thereby preventing the generation of invalid molecules. Figure \ref{fig:smilesselfies} shows the outline of how SMILES and SELFIES represent molecules \cite{20}.

In SMILES, molecules are defined as chains of atoms and represented as strings. Branching within molecules is denoted by parentheses, and pairs of numbers indicate the closure of ring structures. This system can represent complex chemical features such as stereochemistry, aromatic bonds, chirality, ions, and isotopes. 

SELFIES employs a formal grammar-based method where its derivation rules ensure that every combination of symbols corresponds to a chemically valid graph. This attribute effectively prevents the production of invalid molecules, facilitating more efficient compound identification in evolutionary computation. While SELFIES is suitable for representing typical organic molecules, it does not encompass all molecular types.

\subsection{MOEAs}

Single-objective EAs employ a single fitness function to evaluate solutions, therefore, it is trivial to decide whether one solution is better than another or not. In contrast, for MOEAs this is a lot more complicated because each objective needs to be considered. Moreover, there are also instances where two solutions outperform another in differing objectives, meaning that neither is necessarily better than the other. The set that only contains these individuals, which are also better than the rest of the population but not each other, is known as the set of non-dominated solutions or the Pareto-set. The main variation for MOEAs is how the population is maintained and what strategy is applied to obtain said Pareto-set. \\
NSGA-II and NSGA-III both work on two core principles: a ranking technique to highlight favorable solutions and a niching method to increase diversity. Both algorithms employ fast non-dominated sorting to separate the solutions into fronts, based on domination, with the first front containing the non-dominated solutions. Starting with this front, the fronts are added one after another to fill the next population. More often than not, the last front to be included does not exactly fit into the population, thus, niching is used to select the most suitable solutions. For this, NSGA-II uses the crowding distance that estimates the space between a solution and its two neighbors. The rationale behind the crowding distance is that solutions with more space around them are more diverse, hence, better suited. Since calculating the crowding distance in higher dimensions becomes increasingly computationally expensive, NSGA-III uses an alternative approach, utilizing reference directions. These are constructed by uniformly spreading points on the unit simplex and then casting a ray from the origin through each point. To each of these reference directions, solutions are assigned based on closest perpendicular distance. These will then have a higher priority in being chosen for the next population. \\
MOEA/D works by decomposing multi-objective problems into a number of scalar optimization sub-problems, which are then simultaneously optimized. A neighborhood relationship is described, where neighboring sub-problems exchange information and influence another. During the optimization of the sub-problems, the best non-dominated solutions found are preserved in an external population.

\subsection{Genetic Operators and Modifications}

As for the genetic operators, for crossover we make use of one-point crossover between SELFIE strings (Fig. \ref{fig:crossover}). Mutation is designed to add, remove, or replace a single character at an arbitrary position in the SELFIES representation.
For MOEA/D, the offspring replacement behavior must be adjusted to avoid quick convergence to local optima. Typically, a better offspring solution replaces all its neighbours, here we propose to only replace one solution in the neighbourhood. Furthermore, this only happens if its similarity to the neighbourhood does not exceed a certain threshold, to maintain an adequately diverse population.

\subsection{Summary of General Approach}

\begin{figure*}[!htb]
    \centering
    \fbox{\includegraphics[width=0.95\textwidth]{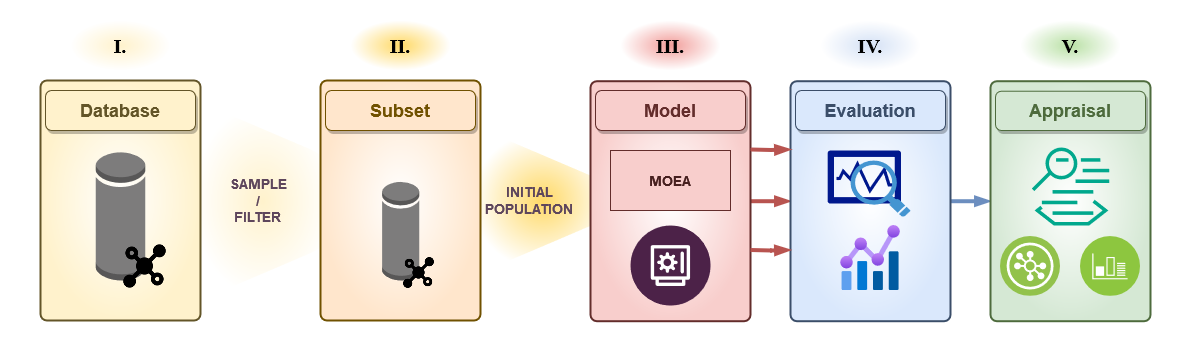}}
    \caption{Proposed CADD pipeline.}
    \label{fig:pipelin}
\end{figure*}

Figure \ref{fig:pipelin} shows the general methodology that we propose in this work. Step one, is to retrieve known compounds that will serve as the starting points of exploration, usually obtainable from publicly available databases. Next, since these databases contain an enormous number of such compounds, we propose to form a reduced subset that also filters out unwanted compounds. This subset can then be used to sample an initial population that is fed into a model of choice to perform drug design (here MOEAs). Afterwards, the model is evaluated for algorithmic performance and overall satisfaction of optimization criteria. Finally, the found solutions are gauged for quality, in our case, this includes whether compounds are novel and possess desired properties.

\section{Experiments}

In the following, we will describe our concrete approach to implementing this pipeline, outlining specific design choices.

\subsection{Subset} \label{subset}
Firstly, for the creation of the subset from which the starting population of compounds will be sampled, we chose the free and commercially available database ZINC \cite{17}. Abiding by Lipinski's Rule of Five \cite{25}, agreed upon rules of thumb for drug-likeness, we downloaded only compounds from ZINC in the range of 250-350 Da of molecular weight and with a LogP value below 5. Since this set still contained over 400 million compounds, we randomly elected 1\%, i.e. around 4 million, of these to keep. Next, we took inspiration from MOSES \cite{31}, a benchmarking suite for molecular generation that also uses a handcrafted subset for its purposes. Applying the filters from MOSES, reduced the final subset to a total of 3.5 million compounds. We further made sure that samples from this subset have aligned properties with the whole subset and that minimal sampling bias was present. Finally, when downloading from ZINC, compounds are encoded already in SMILES format, hence, the last step for the initialization of the populations was to encode the SMILES strings as SELFIES.

\subsection{Objectives}
Our approach is designed to study several objective functions:
\begin{itemize}
    \item The quantitative estimate of drug-likeness (QED) \cite{02}. QED, as the name implies, is a measure to estimate a compounds drug-likeness. A high QED value is obtained if a compound matches the distribution of properties that has been obtained by empirically analyzing known established drugs. In more simple terms, a compound scores well, if it shows similar properties to these known drugs. 
    \item The Synthetic Accessibility score (SA score) \cite{13}, a computational method to gauge the effort required for a compound's synthesis in practice. It ranges from 1 (very simple) to 10 (very difficult). The general idea is to search for complex substructures or features within compounds that are likely to hinder synthesis.
    \item GuacaMol's goal directed multi-property objective (MPO) tasks \cite{06}, containing various objective functions that are specialized to tailor to known marketed drugs. From these, we chose the five tasks Cobimetinib, Fenofexadine, Osimertinib, Pioglitazone, and Ranolazine. A comprehensive list of the MPO objectives is given in \cite{34}.
\end{itemize}

\subsection{Evaluation}

Algorithmic performance assessment is done in three ways: the running metric, internal similarity, and overall inspection of the Pareto-set. The running metric \cite{04} proposed by Blank and Deb, is a metric to rate the performance of MOEAs by analyzing convergence. The broad idea is to move a sliding window over an EAs iterations and studying shifts in the non-dominated solutions. A main advantage of the running metric is that it eliminates the need for the true Pareto-front to be known, as is in our case because we try to discover unseen compounds. Moreover, the running metric can be utilized to directly compare the performance of multiple MOEAs. This is done by accumulating the output Pareto-sets of the different algorithms and using this merged set as a reference to calculate the running metric. 

Another inspiration we took from Verhellen, is to implement an extended similarity index \cite{26, 27}. These are used to measure the internal similarity of a set, resulting in a maximal value if all elements are the same and minimal if all are different. For our purposes, we employ it to measure compound diversity in the Pareto-sets. What is special about this technique is that the calculation is implemented as an n-ary comparison instead of a pairwise one, significantly reducing the computing time. In particular, we used the extended Baroni-Urbani-Buser similarity index. 

\subsection{Compound Appraisal}
One of the major goals of this paper is to analyze the obtained solutions to investigate, if MOEAs can find promising and fully new solutions. This is performed by querying the databases, for whether they contain the obtained Pareto-solutions or not. In case that a Pareto-optimal solution is not in a database, we assume that this could be a new compound, which needs to be further analyzed in terms of desired properties for useful drugs. Besides drug-likeness, there are several other criteria that can be evaluated, using existing analysis tools (e.g., SwissADME \cite{07}).

\subsection{Algorithm Setup}

Each experiment run is setup as follows: for every GuacaMol task, we run all three algorithms for 200 iterations. The population size is either set to 100, or 500, and the reference directions for NSGA-III and MOEA/D are generated using the Riesz energy-based approach \cite{15}. In one such run, the same population sample is reused for comparability. Also, to account for variance, each of these setups is repeated a total of 10 times. For 5 tasks, 2 population sizes, and 10 repetitions, this amounts to $3 * 5 * 2 * 10 = 100$ total experiments. 

To encourage exploration, we set the mutation rate to 80\%, because the algorithms were quick at converging. Also, we normalized the SA score to be in the range of $[0,1]$, since all other objectives behaved the same. 

For both NGSA-II and III, we do not allow duplicates within the population, however, sometimes the Pareto-sets end up with multiple same compounds, as two different SELFIES strings can encode the same compound. 

All experiments were run on a Debian 11 server, with an i7-4790 @3.60GHz CPU, 16GB of DDR3 RAM, and an Intel HD Graphics 4600 GPU.
In terms of software we used the python library Pymoo \cite{03}, a framework for multi-objective optimization that implements the algorithms, as well as the running metric, we deploy.  

\subsection{Results}

\begin{figure}[!htb]
\centering
\begin{subfigure}{0.48\textwidth}
  \centering
  \includegraphics[width=0.49\linewidth]{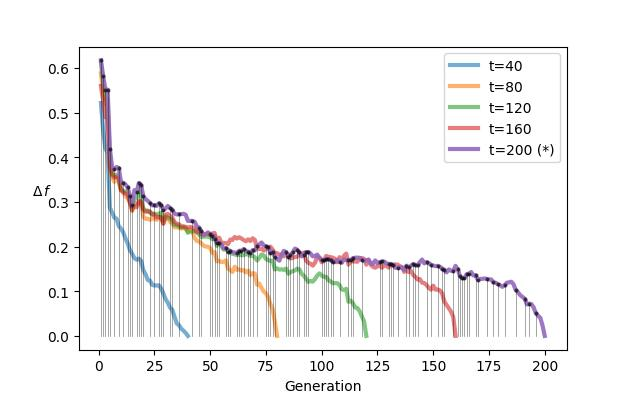}
  \includegraphics[width=0.49\linewidth]{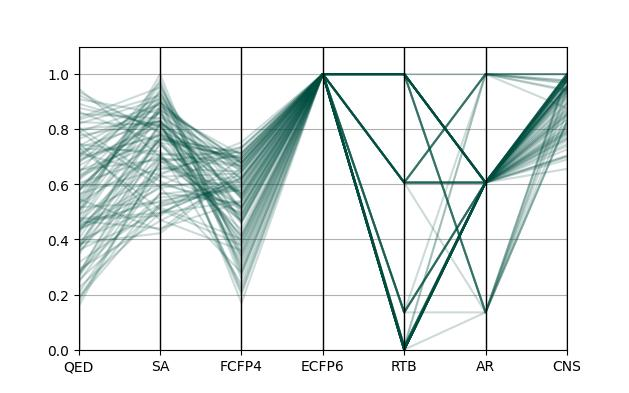}
  \caption{NSGA-II}
\end{subfigure}
\begin{subfigure}{0.48\textwidth}
  \centering
  \includegraphics[width=0.49\linewidth]{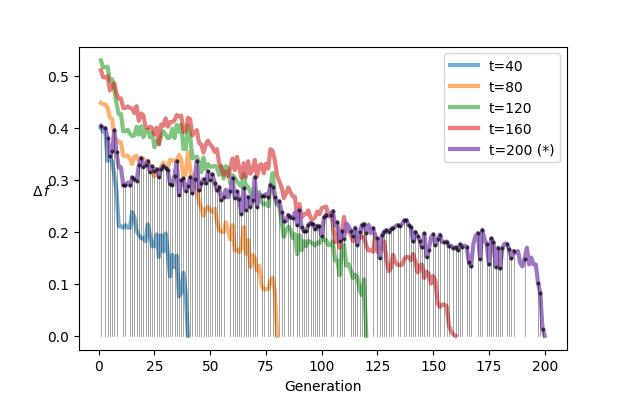}
  \includegraphics[width=0.49\linewidth]{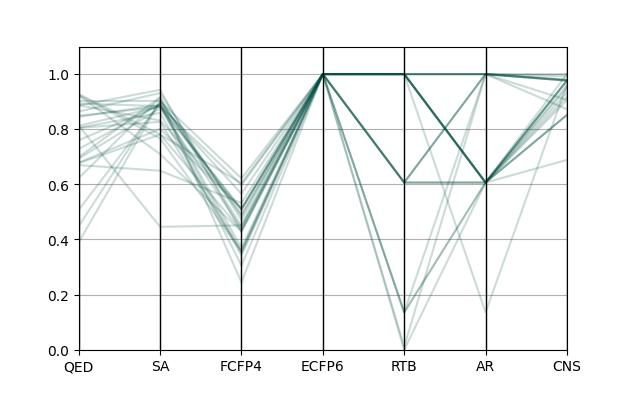}
  \caption{NSGA-III}
\end{subfigure}
\begin{subfigure}{0.48\textwidth}
  \centering
  \includegraphics[width=0.49\linewidth]{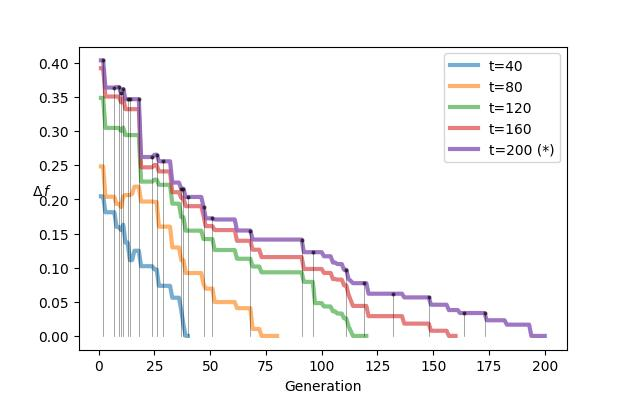}
  \includegraphics[width=0.49\linewidth]{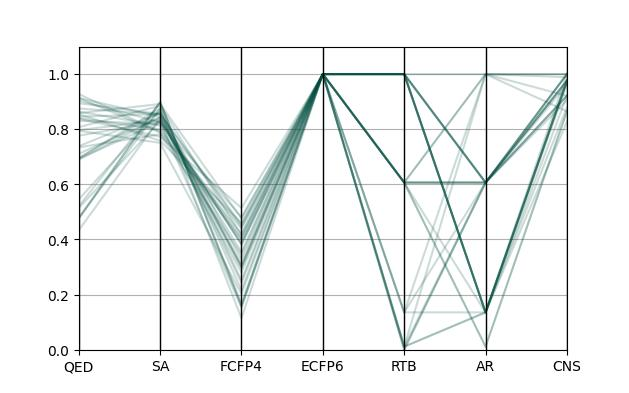}
  \caption{MOEA/D}
\end{subfigure}
\caption{Running Metric and Parallel Coordinate plots - Cobimetinib.}
\label{fig:running_cob}
\end{figure}

To begin with, we will inspect the results of a single experiment run. Fig. \ref{fig:running_cob} shows the running metric and parallel coordinate plots, for each algorithm, for the Cobimetinib task, and a population size of 100. The downwards trajectory in all three running metric graphs, indicates that all algorithms converge, with NSGA-II converging fastest, while NSGA-III and MOEA/D do also, but more gradually.

Adding QED and SA to the five objectives in the Cobimetinib task, yields the displayed total of seven. Overall, NSGA-II finds the most, and thus spread out Pareto-optimal solutions, followed by MOEA/D, and then NSGA-III. A notable observation is that the ECFP6 similarity objective is unnecessary, as it is not conflicting with the other objectives. This was also true for other tasks that included this objective.

Table \ref{tab:obj_cobi} shows the statistics for the objective values as well as the number of Pareto-solutions, per algorithm. Despite the varying numbers of solutions found, the optimal values (MAX columns) do not change much between algorithms. Only for the SA score and the FCFP4 similarity, NSGA-II outperforms the other two. While the averaged values for NSGA-III are higher, this is due to the smaller set size. Compared to NSGA-II, the other two algorithms results are more condensed and skewed towards higher objective values.
 
\begin{table}[t]
\caption{Objective values for Cobimetinib task.}
\resizebox{\columnwidth}{!}{%
\begin{tabular}{@{}lccccccccc@{}}
\toprule
\multicolumn{1}{c}{\multirow{2}{*}{\textbf{Objectives}}} & \multicolumn{3}{c}{\textbf{NSGA-II}}                                        & \multicolumn{3}{c}{\textbf{NSGA-III}}                                       & \multicolumn{3}{c}{\textbf{MOEA/D}}                                         \\
\multicolumn{1}{c}{}                                     & \multicolumn{1}{l}{MIN} & \multicolumn{1}{l}{MAX} & \multicolumn{1}{l}{AVG} & \multicolumn{1}{l}{MIN} & \multicolumn{1}{l}{MAX} & \multicolumn{1}{l}{AVG} & \multicolumn{1}{l}{MIN} & \multicolumn{1}{l}{MAX} & \multicolumn{1}{l}{AVG} \\ \midrule
QED                                                      & 0.16                    & \textbf{0.95}                    & 0.55                    & 0.39                    & 0.93                    & 0.76                    & 0.43                    & 0.93                    & 0.74                    \\
SA                                                       & 0.42                    & \textbf{1}                       & 0.74                    & 0.45                    & 0.94                    & 0.84                    & 0.75                    & 0.9                     & 0.84                    \\
FCFP4                                                    & 0.17                    & \textbf{0.75}                    & 0.53                    & 0.24                    & 0.62                    & 0.45                    & 0.11                    & 0.51                    & 0.33                    \\
ECFP6                                                    & 1                       & \textbf{1}                       & 1                       & 1                       & \textbf{1}                       & 1                       & 1                       & \textbf{1}                       & 1                       \\
RTB                                                      & 0                       & \textbf{1}                       & 0.44                    & 0                       & \textbf{1}                       & 0.72                    & 0                       & \textbf{1}                       & 0.61                    \\
AR                                                       & 0.14                    & \textbf{1}                       & 0.6                     & 0.14                    & \textbf{1}                       & 0.75                    & 0.01                    & \textbf{1}                       & 0.46                    \\
CNS                                                      & 0.66                    & \textbf{1}                       & 0.91                    & 0.69                    & \textbf{1}                       & 0.94                    & 0.83                    & \textbf{1}                       & 0.95                    \\
\#Pareto                                                 & \multicolumn{3}{c}{\textbf{100}}                                                     & \multicolumn{3}{c}{25}                                                      & \multicolumn{3}{c}{31}                                                      \\ \bottomrule
\end{tabular}
}

\label{tab:obj_cobi}
\end{table}

\begin{figure*}[!htb]
\centering  
\begin{subfigure}{0.25\textwidth}
  \includegraphics[width=\linewidth]{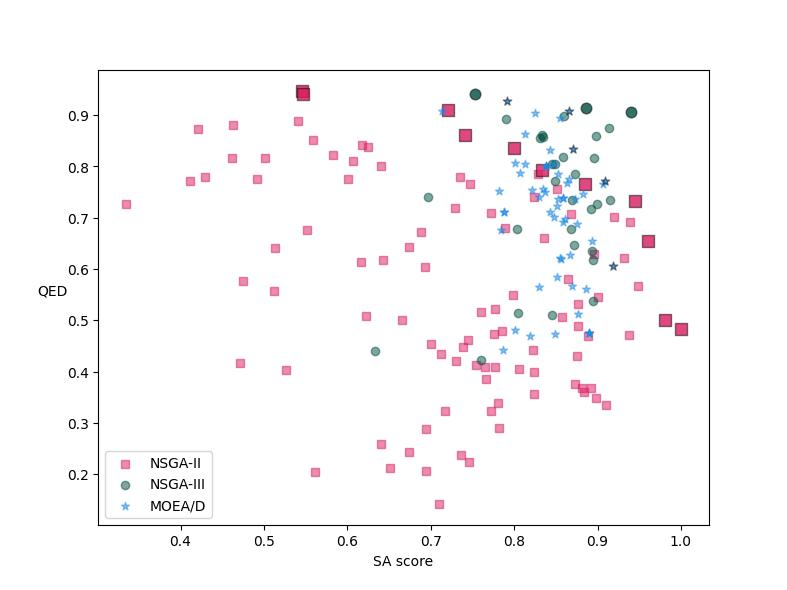}
  \caption{Population size 100}
  \label{fig:pareto_cob_100}
\end{subfigure}
\begin{subfigure}{0.25\textwidth}%
  \includegraphics[width=\linewidth]{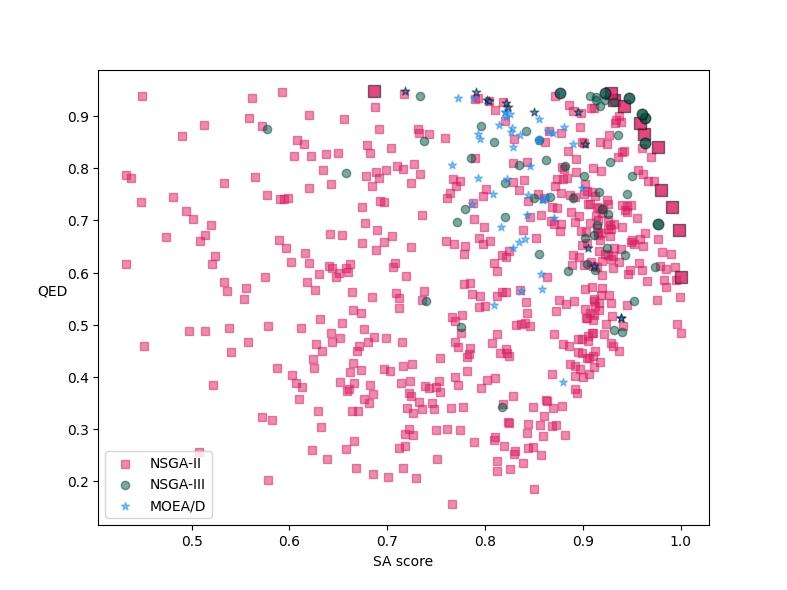}
  \caption{Population size 500}
  \label{fig:pareto_cob_500}
\end{subfigure}
\begin{subfigure}{0.25\textwidth}%
  \fbox{\includegraphics[width=\linewidth]{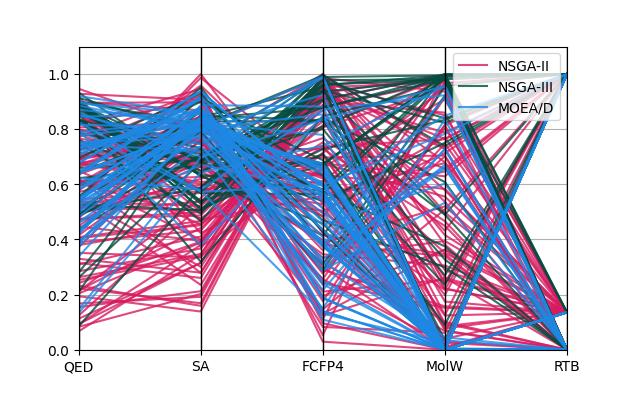}}
    \caption{PC-plot Pioglitazone}
    \label{fig:pc_piog}
\end{subfigure}
\caption{Pareto-plots for Cobimetinib and PC-plot for Pioglitazone.}
\end{figure*}

This is more apparent upon closer inspection of the Pareto-sets. Fig. \ref{fig:pareto_cob_100} shows the three resulting sets plotted for QED and SA score. As visible, compounds NSGA-II found are spread out further over the search space, whereas NSGA-III and MOEA/D find fewer solutions that are in a closer vicinity to another. Nevertheless, there are only few overlaps between the sets, meaning that all algorithms produce Pareto-optimal solutions, i.e. potential compounds.

Similar outcomes are obtained when the population size is increased to 500 (see fig. \ref{fig:pareto_cob_500}). Although with increased population size, NSGA-III finds more non-dominated solutions than MOEA/D.

A task where all algorithms performed well, is the Pioglitazone task with five total objectives. Fig. \ref{fig:pc_piog} displays the parallel coordinate plot with all algorithm results overlaid, and table \ref{tab:obj_piog} the objective statistics. NSGA-II still attains the
largest Pareto-set but NSGA-III and MOEA/D fill about half of the population with non-dominated individuals. All algorithms reach close to ideal objective values with minor differences.

\begin{table}[t]
\caption{Objective values for Pioglitazone task.}
\resizebox{\columnwidth}{!}{%
\begin{tabular}{@{}lccccccccc@{}}
\toprule
\multicolumn{1}{c}{\multirow{2}{*}{\textbf{Objectives}}} & \multicolumn{3}{c}{\textbf{NSGA-II}}                                        & \multicolumn{3}{c}{\textbf{NSGA-III}}                                       & \multicolumn{3}{c}{\textbf{MOEA/D}}                                         \\
\multicolumn{1}{c}{}                                     & \multicolumn{1}{l}{MIN} & \multicolumn{1}{l}{MAX} & \multicolumn{1}{l}{AVG} & \multicolumn{1}{l}{MIN} & \multicolumn{1}{l}{MAX} & \multicolumn{1}{l}{AVG} & \multicolumn{1}{l}{MIN} & \multicolumn{1}{l}{MAX} & \multicolumn{1}{l}{AVG} \\ \midrule
QED                                                      & 0.05                    & \textbf{0.95}                    & 0.53                    & 0.15                    & 0.92                    & 0.68                    & 0.31                    & 0.91                    & 0.65                    \\
SA                                                       & 0.11                    & \textbf{1}                       & 0.62                    & 0.42                    & \textbf{1}                       & 0.73                    & 0.58                    & 0.92                    & 0.81                    \\
FCFP4                                                    & 0.02                    & \textbf{1}                       & 0.66                    & 0.06                    & \textbf{1}                       & 0.69                    & 0.07                    & 0.98                    & 0.47                    \\
MolW                                                     & 0                       & \textbf{1}                       & 0.46                    & 0                       & \textbf{1}                       & 0.58                    & 0                       & \textbf{1}                       & 0.24                    \\
RTB                                                      & 0                       & \textbf{1}                       & 0.27                    & 0                       & \textbf{1}                       & 0.54                    & 0                       & \textbf{1}                       & 0.17                    \\
\#Pareto                                                 & \multicolumn{3}{c}{\textbf{100}}                                                     & \multicolumn{3}{c}{44}                                                      & \multicolumn{3}{c}{53}                                                      \\ \bottomrule
\end{tabular}
}

\label{tab:obj_piog}
\end{table}

\begin{figure}[!htb]
\centering
\begin{subfigure}{0.49\columnwidth}
  \centering
  \includegraphics[width=\linewidth]{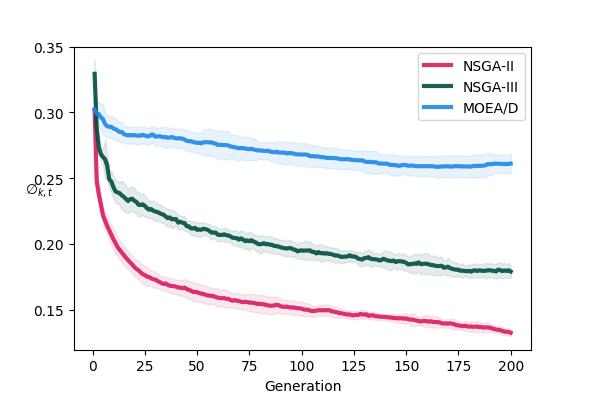}
  \caption{Pioglitazone} \label{fig:run_cobi}
\end{subfigure} \hfill
\begin{subfigure}{0.49\columnwidth}
  \centering
  \includegraphics[width=\linewidth]{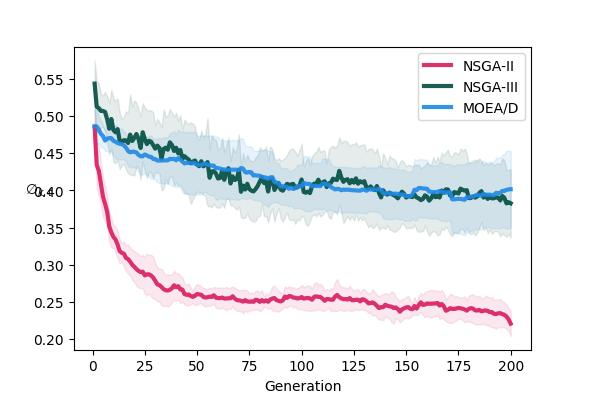}
  \caption{Cobimetinib} \label{fig:run_piog}
\end{subfigure}
\begin{subfigure}{0.49\columnwidth}
  \centering
  \includegraphics[width=\linewidth]{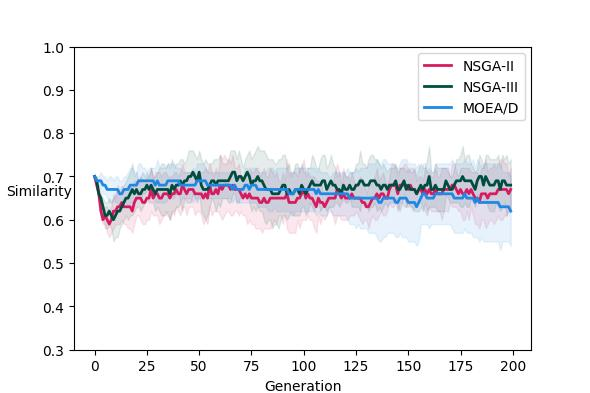}
  \caption{Fenofexadine} \label{fig:sim_fex}
\end{subfigure} \hfill
\begin{subfigure}{0.49\columnwidth}
  \centering
  \includegraphics[width=\linewidth]{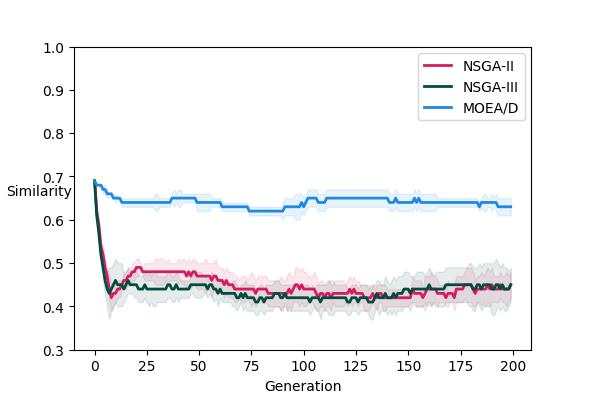}
  \caption{Pioglitazone} \label{fig:sim_piog}
\end{subfigure}
\caption{Running Metric and Internal Similarity.}
\end{figure}

Moving on to the accumulated results of 10 runs, we will first contrast the results for the Cobimetinib task. Table \ref{tab:obj_cobi_avg} shows the objective statistics averaged over 10 runs, along with the standard deviation. With low variance between runs, the objective values and the Pareto-set sizes match the previous observations in the single run. NSGA-II always manages to discover a full population of non-dominated solutions, regardless of population size. Whilst MOEA/D found more compounds than NSGA-III in most 100 population size runs, interestingly, when the population size is increased to 500, NSGA-III finds more non-dominated solutions in most cases. \\
For instance in the Pioglitazone task, NSGA-III finds 145 Pareto-solutions, as opposed to MOEA/D's 128, and also outperforms MOEA/D in terms of objective values. \\
This is more discernible in the running metric analysis. To reiterate, the running metric works by merging the final Pareto-sets obtained by each algorithm and uses that set as a reference to measure performance. Fig. \ref{fig:run_cobi} displays the corresponding plot, clearly showing that NSGA-III outperforms MOEA/D. 
This was consistent across all experiments with a population size of 500, even when MOEA/D found more solutions, indicating that the quality of compounds found by NSGA-III are better. Going back to a population size of 100, there are however also cases were MOEA/D competes well with NSGA-III, as can be seen in Fig. \ref{fig:run_piog}. NSGA-II, always outperforms the other two algorithms, which is to be expected, since the objective values in NSGA-II are high and the contribution to the reference set is largest.

\begin{table*}[t]
\caption{Averaged objective values over 10 runs for Cobimetinib task.} 
\resizebox{\textwidth}{!}{%
\begin{tabular}{@{}lccccccccc@{}}
\toprule
\multicolumn{1}{c}{\multirow{2}{*}{\textbf{Objectives}}} & \multicolumn{3}{c}{\textbf{NSGA-II}}                                        & \multicolumn{3}{c}{\textbf{NSGA-III}}                                       & \multicolumn{3}{c}{\textbf{MOEA/D}}                                         \\
\multicolumn{1}{c}{}                                     & \multicolumn{1}{l}{MIN} & \multicolumn{1}{l}{MAX} & \multicolumn{1}{l}{AVG} & \multicolumn{1}{l}{MIN} & \multicolumn{1}{l}{MAX} & \multicolumn{1}{l}{AVG} & \multicolumn{1}{l}{MIN} & \multicolumn{1}{l}{MAX} & \multicolumn{1}{l}{AVG} \\ \midrule
QED                                                      & \textbf{0.15} \tiny{± 0.03}    & \textbf{0.95} \tiny{± 0.0}     & \textbf{0.56} \tiny{± 0.03}    & \textbf{0.45} \tiny{± 0.08}    & \textbf{0.94} \tiny{± 0.01}    & \textbf{0.76} \tiny{± 0.02}    & \textbf{0.43} \tiny{± 0.04}    & \textbf{0.93} \tiny{± 0.01}    & \textbf{0.73} \tiny{± 0.02}    \\
SA                                                       & \textbf{0.31} \tiny{± 0.07}    & \textbf{1.0} \tiny{± 0.0}      & \textbf{0.72} \tiny{± 0.02}    & \textbf{0.54} \tiny{± 0.11}    & \textbf{0.94} \tiny{± 0.01}    & \textbf{0.82} \tiny{± 0.02}    & \textbf{0.73} \tiny{± 0.05}    & \textbf{0.94} \tiny{± 0.04}    & \textbf{0.85} \tiny{± 0.01}    \\
FCFP4                                                    & \textbf{0.14} \tiny{± 0.03}    & \textbf{0.77} \tiny{± 0.04}    & \textbf{0.5} \tiny{± 0.03}     & \textbf{0.26} \tiny{± 0.02}    & \textbf{0.66} \tiny{± 0.04}    & \textbf{0.47} \tiny{± 0.02}    & \textbf{0.17} \tiny{± 0.06}    & \textbf{0.52} \tiny{± 0.03}    & \textbf{0.36} \tiny{± 0.02}    \\
ECFP6                                                    & \textbf{1.0} \tiny{± 0.0}      & \textbf{1.0} \tiny{± 0.0}      & \textbf{1.0} \tiny{± 0.0}      & \textbf{1.0} \tiny{± 0.0}      & \textbf{1.0} \tiny{± 0.0}      & \textbf{1.0} \tiny{± 0.0}      & \textbf{1.0} \tiny{± 0.0}      & \textbf{1.0} \tiny{± 0.0}      & \textbf{1.0} \tiny{± 0.0}      \\
RTB                                                      & \textbf{0.0} \tiny{± 0.0}      & \textbf{1.0} \tiny{± 0.0}      & \textbf{0.47} \tiny{± 0.05}    & \textbf{0.03} \tiny{± 0.05}    & \textbf{1.0} \tiny{± 0.0}      & \textbf{0.75} \tiny{± 0.04}    & \textbf{0.0} \tiny{± 0.01}     & \textbf{1.0} \tiny{± 0.0}      & \textbf{0.7} \tiny{± 0.08}     \\
AR                                                       & \textbf{0.06} \tiny{± 0.06}    & \textbf{1.0} \tiny{± 0.0}      & \textbf{0.6} \tiny{± 0.05}     & \textbf{0.23} \tiny{± 0.19}    & \textbf{1.0} \tiny{± 0.0}      & \textbf{0.73} \tiny{± 0.05}    & \textbf{0.09} \tiny{± 0.06}    & \textbf{1.0} \tiny{± 0.0}      & \textbf{0.52} \tiny{± 0.08}    \\
CNS                                                      & \textbf{0.58} \tiny{± 0.07}    & \textbf{1.0} \tiny{± 0.0}      & \textbf{0.88} \tiny{± 0.02}    & \textbf{0.83} \tiny{± 0.05}    & \textbf{1.0} \tiny{± 0.0}      & \textbf{0.95} \tiny{± 0.01}    & \textbf{0.85} \tiny{± 0.03}    & \textbf{1.0} \tiny{± 0.0}      & \textbf{0.96} \tiny{± 0.01}    \\
\#Pareto                                                 & \multicolumn{3}{c}{\textbf{100.0} \tiny{± 0.0}}                                    & \multicolumn{3}{c}{\textbf{27.0} \tiny{± 2.0}}                                     & \multicolumn{3}{c}{\textbf{39.0} \tiny{± 8.0}}                                     \\ \bottomrule
\end{tabular}
}

\label{tab:obj_cobi_avg}
\end{table*}

Lastly, the internal similarity plotted over the generations are depicted in figures \ref{fig:sim_fex} and \ref{fig:sim_piog}. In most experiments the similarity was stable and it hovered around 0.6 to 0.7. The only notable differing case is the Pioglitazone task, where the internal similarities for NSGA-II and III decrease to 0.4 and 0.5, well below MOEA/D. These results indicate that the algorithms are able to maintain sufficient diversity and do not suffer from negative genetic drift.

\subsection{Final check}

\begin{figure}[!ht]
\centerline{\includegraphics[width=0.4\textwidth]{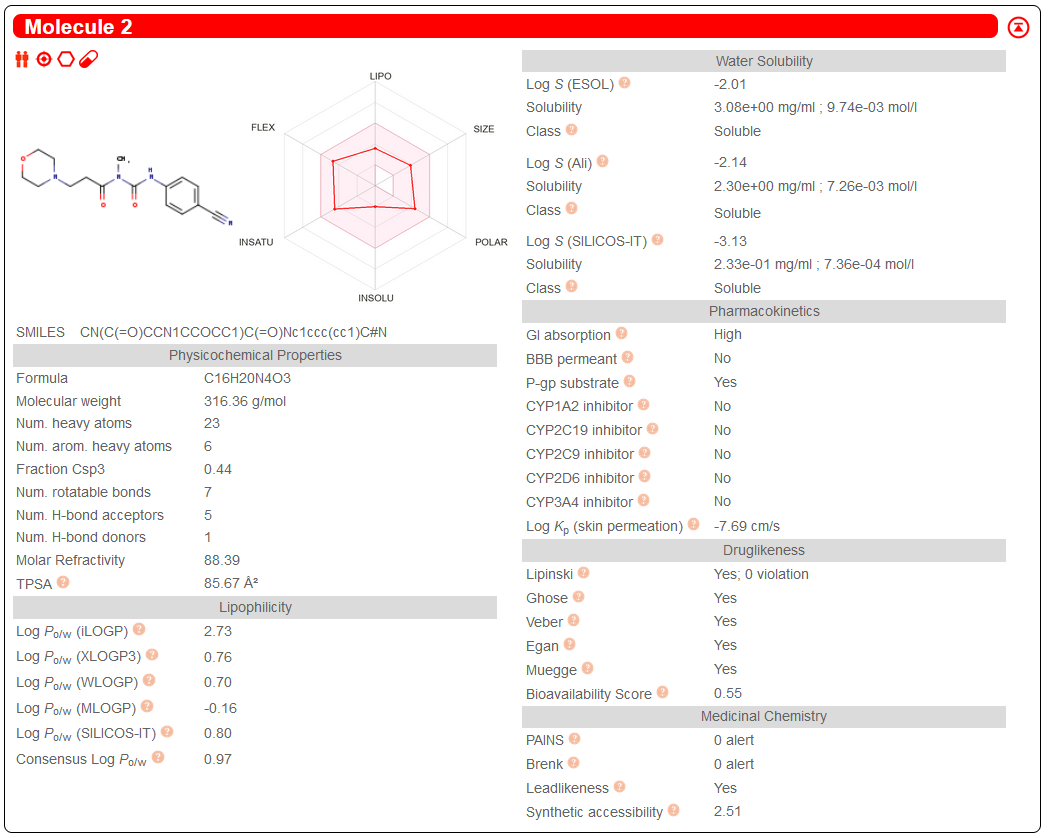}}
\caption{SwissADME output.}
\label{fig:swiss}
\end{figure}

\begin{figure*}[!htb]
\centerline{\includegraphics[width=0.9\textwidth]{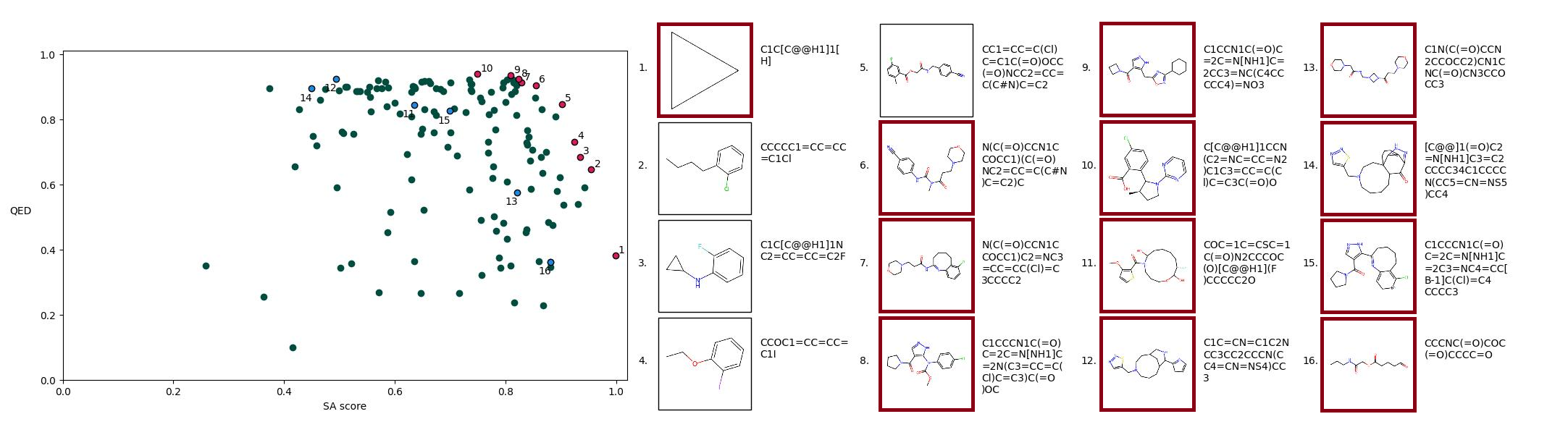}}
\caption{Pareto-plot plus Molecular Graphs}
\label{fig:pareto_mols}
\end{figure*}

For the assessment of the produced solutions, the Pareto-sets were again plotted for QED and SA score, and non-dominated solutions in this space were emphasized along with their corresponding molecular graphs (see fig. \ref{fig:pareto_mols}). In case there were only few non-dominated solutions, we randomly picked more compounds to visualize. The highlighted molecular graphs are compounds that returned no match when queried against the ZINC database, suggesting they are novel. These were then further analyzed using the SwissADME tool. Its output for an exemplary solution candidate can be seen in fig. \ref{fig:swiss}. The red colored area in the radar chart indicates the space of suitable compounds, in which, our newly discovered compound resides in. Furthermore, it satisfies all drug-likeness metrics and bypasses PAINS \cite{01} and Brenk filters \cite{05} that detect problematic substructures. We manually checked a handful of compounds this way and found that they are generally of decent quality, regarding SwissADME, but not without a few imperfections.

\subsection{Discussion}

In summary, the major disparity between the algorithms is the number of Pareto-optimal solutions, i.e. potential compounds, found. NSGA-II performed best for all tasks and its final population consisted always of only non-dominated solutions, regardless of population size. MOEA/D had competing outcomes with NSGA-III for smaller population sizes, but especially for larger population sizes, NSGA-III produced better results. Still, the analysis of the resulting compounds revealed that among the solutions, novel compounds with desirable properties, for all algorithms, are discovered. This implies that in principle all algorithms function successfully.

\section{Conclusion}

In this paper, we introduce a CADD approach for the design of new drug candidate compounds, employing the string representation method SELFIES in conjunction with MOEAs. Our approach involves generating a subset of promising compounds and sampling from this subset to form an advantageous initial population. The exploration of the chemical space is conducted using crossover and mutation on the SELFIES representation, which does not produce invalid offspring and, therefore, does not require a repair mechanism.

We utilized the ZINC database in our numerical experiments and applied several MOEAs, including NSGA-II, NSGA-III, and MOEA/D. The optimization criteria were based on QED, SA score, and objectives from the GuacaMol benchmark tasks. Convergence and diversity were measured using execution metrics and internal similarity, respectively. Furthermore, the novelty and desirable properties of the obtained solutions were evaluated. The results demonstrated that all algorithms successfully generated promising compounds, with NSGA-II discovering the most novel compounds and exhibiting the best performance.

The main advantage of our EA-based approach, as often emphasized in the literature \cite{18, 16, 28}, lies in its overall simplicity and ease of interpretation. Additionally, the computational efficiency of the EA is demonstrated, with experiments requiring a day of execution time at most.


A forthright improvement to our method is to analyze the candidate compounds further. This procedure was done manually in our pipeline on only a handful of solutions, hence, automating this process is a priority. Also, the filters we applied to our initial sampling set could be helpful if administered in the latter steps of the pipeline. This improvement would give a more precise impression of how many potential compounds are found and if chemical space is being sufficiently explored. Finally, adjustments to our evolutionary approach are of interest to avoid fast convergence. For instance, replacing the one-point crossover with a more sophisticated method will increase the diversity of offspring solutions.

\section*{Acknowledgment}
This work is partially funded by the German Federal Ministry of Education and Research through the 6G-ANNA project (grant no. 16KISK092).

\renewcommand*{\bibfont}{\footnotesize}

\end{document}